\documentclass[10pt, a4paper, conference]{IEEEtran}

\usepackage[T1]{fontenc}

\usepackage{url}

\usepackage{graphicx}
\usepackage[cmex10]{amsmath}
\usepackage{amsfonts}

\begin{document}

\title{Down-Sampling coupled to Elastic Kernel Machines for Efficient Recognition of Isolated Gestures}

\author{\IEEEauthorblockN{Pierre-Francois Marteau and Sylvie Gibet and Cl\'{e}ment Reverdy}
\IEEEauthorblockA{IRISA (UMR 6074),  Universit\'{e} de Bretagne Sud\\
Campus de Tohannic, 56000 Vannes, France\\
Email: firstname.name AT univ-ubs DOT fr}
}

\maketitle

\begin{abstract}
In the field of gestural action recognition, many studies have focused on dimensionality reduction along the spatial axis, to reduce both the variability of gestural sequences expressed in the reduced space, and the computational complexity of their processing. It is noticeable that very few of these methods have explicitly addressed the dimensionality reduction along the time axis. This is however a major issue with regard to the use of elastic distances characterized by a quadratic complexity. To partially fill this apparent gap, we present in this paper an approach based on temporal down-sampling associated to elastic kernel machine learning. 
We experimentally show, on two data sets that are widely referenced in the domain of human gesture recognition, and very different in terms of quality of motion capture, that it is possible to significantly reduce the number of skeleton frames while maintaining a good recognition rate. The method proves to give satisfactory results at a level currently reached by state-of-the-art methods on these data sets. The computational complexity reduction makes this approach eligible for \textit{real-time} applications.
\end{abstract}

\section{Introduction}

During the past decade, gesture recognition has been a very active research field that has evolved in terms of improving motion capture technology and recognition methods, mostly based on machine learning techniques. Recently, the availability of low-cost consumer technology, often associated with game consoles, has helped to democratize the use of motion sensors, not only in the context of interactive video game, but also in various frameworks using gestural interaction. 
Hence high quality databases, built from expensive motion capture (\emph{mocap}) devices requiring specific expertise, exist today alongside lower quality databases, i.e. containing noisier and less accurate data provided by new inexpensive sensors that require very little expertise. Therefore, heterogeneous databases of captured motion of various qualities are available to the scientific community, and comparing the robustness and generalization of recognition algorithms on these diverse motion databases is highly challenging.
Beyond the quality of the recognition, the complexity of the algorithms and their response time is also a major issue, especially for real-time interaction.

In the context of the recognition of isolated gestures from motion captured data, we present in this paper a novel method that improves the performance of classical support vector machines when used with regularized elastic kernels. We also show how the temporal dimensionality reduction, associated with such elastic kernels significantly improves the efficiency of the algorithm and the recognition scores. 

\section{Motion captured data and sequence of skeletal poses}
We focus on human motion data captured by various camera-based sensors (infrared marker-tracking system with high resolution, or webcam-style system). The data is uniformly preprocessed so that the captured data is finally reconstructed as a set of 3D-trajectories of the skeleton joints determined from the positions of markers captured on a real actor underlying more or less accurately the skeleton of the actor who produced the movement. 
The identification of the skeleton model from captured data is achieved through a mapping-optimization process such as the ones described in \cite{AguiarTS06}, \cite{Obrien:2000}, or \cite{Shotton:2011}. The techniques based on a skeleton model hence convert 3D sensor data into Cartesian or angular coordinates that define with various accuracies the state of the joints over time.

\begin{figure*}[ht!]
  \centering
  \begin{tabular}{cc}
	\includegraphics[width=70mm]{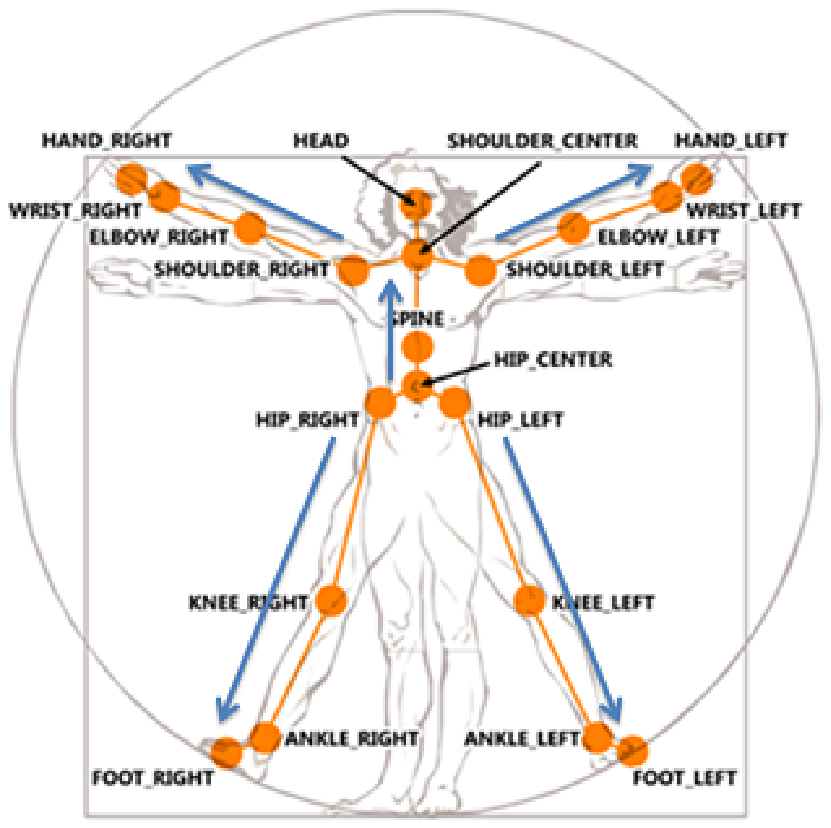}&
    \includegraphics[width=50mm]{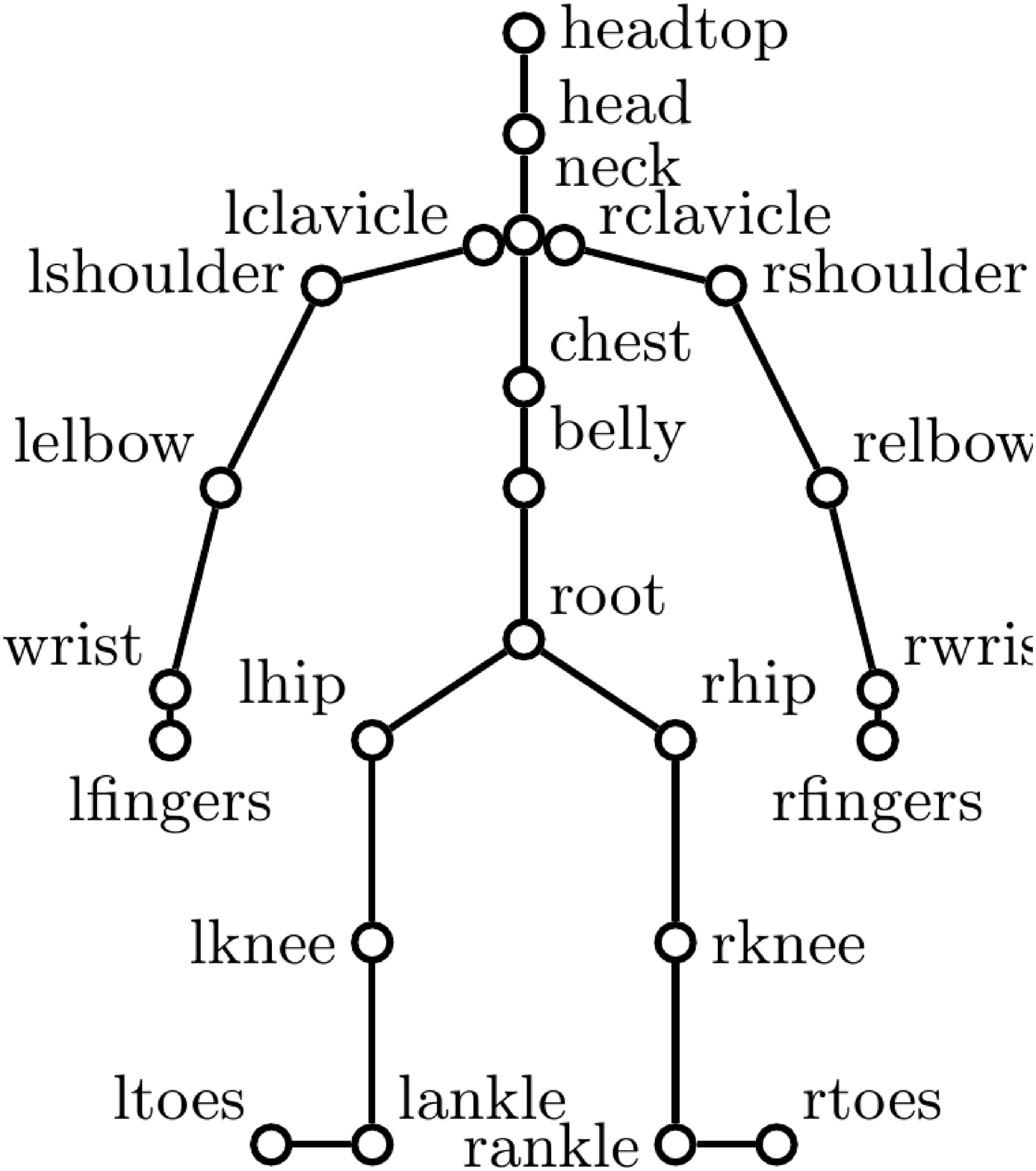}
  \end{tabular}
  \caption{Examples of skeletons reconstructed from motion data captured from the Kinect (left) and from the Vicon-MX device used by the Max Planck Institute (right).}
\label{fig:skels}
\end{figure*}

Figure \ref{fig:skels} presents two skeletons reconstructed from two very distinct capture systems. On the right, the skeleton is reconstructed from data acquired via the Microsoft Kinect, on the left from the Vicon-MX device used by the Max Planck Institute to produce the  HDM05 datasets.

This sort of data is inherently noisy, mainly due to the data acquisition process (drifts, imprecision and shading, etc.) and  sensor noise. Furthermore, due to the nature of the capture devices and in particular to the number of sensors that are operated, the nature of the reconstructed skeletons is also subject to  some variation for two main reasons: in the one hand the morphology of the actor (segment lengths) is the main source of variability for a given capture device, and in the second hand, the number of joints (the number of degree of freedom) vary according to the capture device, leading to some additional noise due to the reconstruction of the skeletal data from sensor outputs.

Thus, any movement can be defined as a multivariate state vector describing a trajectory over time, i.e. a time series
$\{Y_t \in\mathbf{R}^{k}\}^{T}_{1} = [Y_1 , . . . , Y_T]$, where the $k$ spatial dimension ($k=3\cdot N$, with N the number of joints) typically varies between $20$ and $100$ according to the capture devices and the considered task. 
As this state vector is obviously not composed of independent scalar dimensions, the spatio-temporal encoded redundancies open new prospects for dimension reduction as well as noise reduction approaches, which is particularly relevant when considering motion recognition as one can targets gains both in terms of computation time and error rate.

\section{Related works}
Gesture analysis and recognition is very broad and recently very active area due to the democratization of low cost motion capture systems by camera. It covers aspects of signal and image processing, dynamic modeling, statistical and machine learning approaches. 
We hereinafter give a brief non-exhaustive overview of the main methods proposed for gesture analysis and recognition. These methods primarily focus on the extraction of features that significantly represent the kinematics and dynamics of the skeleton data characterized as a whole, or portions of it. 
Among them, some approaches, based on linear dynamic models  \cite{Veeraraghavan2004}, have exploited autoregressive (AR) models and autoregressive moving-average (ARMA) models to characterize the kinematics of movements, while other approaches, based on nonlinear dynamic models \cite{Bissacco2007}, have developed movement analysis and recognition scheme based on dynamical models controlled by Gaussian processes. \cite{Mitra:2007} propose a synthesis of the major gesture recognition approaches relying on Hidden Markov Models (HMM).
\cite{Wang2006} have exploited conditional random fields to model joint dependencies and thus increase the discrimination of HMM-like models. Recurrent neural network models have also been used \cite{Martens2011}; among them, conditional restricted Boltzman's machines \cite{Larochelle2012} have been studied in the context of motion captured data modeling and classification.

Some methods address more specifically the problem of dimensionality reduction with an objective for reducing the variability while seeking an efficiency gain. 
In particular, the Principal Component Analysis (PCA) has been widely exploited for gesture analysis and recognition with the objective of reducing the dimensionality of motion data \cite{Masoud2003}. Other methods such as linear projections preserving locally neighborhoods (Locality Preserving Projection) \cite{He2003LPP}, or their non-linear counter-parts such as ISOMAP \cite{Tenenbaum2000}, have been implemented to embed postures in low dimensional spaces in which a more efficient time warp (DTW, see section \ref{sec:DTW}) algorithm, associated with the Hausdorff distance, can be used to classify movements.
Other ad hoc methods for dimension reduction are also proving their efficiency: we can mention for example the recent work of \cite{YuAggarwal2009} that proposes to only consider the  trajectories of the 5 end-extremities of the skeleton (2 feet, 2 hands and the head).
Models based on Gaussian processes with latent variables are also largely used, for instance a hierarchical version has been recently exploited for gesture recognition \cite{Han:2010}.

The identification of significant variables maximizing the discrimination of motion classes has also been widely explored. \cite{Fothergill:2012} and \cite{Zhao2012} have in particular applied random forests to recognize actions, using a Kinect sensor, while \cite{OfliF2013} recently proposed to automatically select the most informative skeletal joints to explain the current action.
In the same line, \cite{Hussein2013} consider covariance matrices evaluated on some skeletal joints as discriminative descriptors to characterize a movement sequence. The use of sliding windows can be view as a dimension reduction along the time axis. In \cite{Li2010} a simple bag of 3D points is used to represent and recognize gestural action. Similarly, in \cite{Wang2012}, \textit{actionlets} are defined from Fourrier's coefficients to characterize the most discriminative joints. Histograms of oriented 4D normals have been also proposed in \cite{Oreifej2013} for the recognition of gestural actions and movements from sequences of depth images. 

Finally, it can be mentioned, among many existing applications that address the use of elastic distances into a recognition process, the recent work described in \cite{Sempena2011},  as well as the hardware acceleration proposed in \cite{Hussain2012}. However, to our knowledge, no work exploiting this type of distance has directly studied the question of data reduction along the time axis.

\section{Downsampling movement sequences coupled to elastic kernel machines}
When considering the use of elastic distances or kernels to benefit from their ability to deal with some form of temporal variability, we are rapidly confronted with their computational cost, in general quadratic with the length of the time series that are processed and linear with the \textit{spatial} dimension (number of degrees-of-freedom). 
This high computational complexity is somehow limiting their use, especially when large amounts of data has to be processed, or when so-called \textit{real-time} constraint is required. It is therefore \textit{a priori} particularly relevant to consider a dual dimensionality reduction, firstly on the time axis, and secondly on the spatial axis. 
Hence, in the context of \textit{mocap} data processing, it seems useful to determine if a spatio-temporal redundancy of motion paths can be exploited, especially in the perspective of using elastic distances.

\subsection{Dimension reduction along the time axis}
Considering the quite rich literature on gesture recognition, it is significant to note that while some studies have shown success with dimensionality reduction on the spatial axis, very few have directly addressed a reduction in dimensionality along the time axis \textit{per se}  to reduce the complexity of elastic matching.
\cite{Keogh:2000} explicitly mentioned a temporal sub-sampling associated with a dynamic time warping in the context of time series mining, followed later by \cite{MarteauTWED09}.
In order to explicitly reduce dimensionality along the time axis, our straightforward approach here consists in sub-sampling the motion data so that each motion trajectory takes the form of a fixed-size sequence of $L$ skeletal postures, evenly distributed along the time axis. 
It becomes then easy to perform a classification or recognition task by using elastic kernel machines on such fixed-size sequences to assess performance rates depending on the degree of sub-sampling that is considered. Indeed, this approach is quite raw, as long sequences can be characterized with the same number of skeletal poses than short sequences. For very short sequences, whose length is shorter than $L$, if any, we  over-sample the sequence in order to meet the fixed-size requirement. But we consider this case as very marginal here since we seek a sub-sampling rate much lower than the average length of the motion sequence. 
\subsection{Elastic kernels and their regularization}    
\label{sec:DTW}
\textbf{Dynamic Time Warping} (DTW), \cite{TWED:Velichko70}, \cite{TWED:Sakoe71}, by far the most used elastic measure, is defined as 
\begin{eqnarray}
\label{Eq.2}
 d_{dtw}(X_p,Y_q)&= &d_{E}^{2}(x(p),y(q))  \\
  &+&\text{Min} 
   \left\{
   \begin{array}{ll}
     d_{dtw}(X_{p-1},Y_{q}) & sup \nonumber\\ 
     d_{dtw}(X_{p-1},Y_{q-1}) & sub	\nonumber	 \\
     d_{dtw}(X_{p},Y_{q-1}) & ins \nonumber \\
   \end{array}
   \right.
\end{eqnarray}
where $d_{E}(x(p),y(q)$ is the Euclidean distance (possibly the square of the Euclidean distance) defined on $\mathbb{R}^k$ between the two postures in sequences $X$ and $Y$ taken at times $p$ and $q$ respectively. Besides the fact that this measure does not respect the triangle inequality, it does not directly define a positive definite kernel. 
When performed by a support vector machine (SVM) model, the optimization problem inherent to this type of learning algorithm is no longer quadratic. 
Moreover, the convergence towards the optimorum is no longer guaranteed, which, depending on the complexity of the task may be considered as detrimental.
Besides the fact that the DTW measure does not respect the triangle inequality, it is furthermore not possible to directly define a positive definite kernel from it. Hence, the optimization problem, inherent to the learning of a kernel machine, is no longer quadratic which could, at least on some tasks, be a source of limitation.\\

\textbf{Regularized DTW}: recent works  \cite{Cuturi07},  \cite{Marteau2014} allowed to propose new guidelines to regularize kernels constructed from elastic measures such as DTW. A simple instance of such regularized kernel, derived from \cite{Marteau2014} for time series of equal length, takes the following form, which relies on two recursive terms :

\begin{align}
\label{Eq.MEREDK}
\begin{array}{ll}
\mathcal{K}_{rdtw}(X_{p},Y_{q})=K^{xy}_{rdtw}(X_{p}, Y_{q})+K^{xx}_{rdtw}(X_{p},Y_{q}) \\
\\
K^{xy}_{rdtw}(X_{p},Y_{q}) = \frac{1}{3}e^{-\nu d_{E}^{2}(x(p),y(q))}  \\
   \sum \left\{
   \begin{array}{ll}
    h(p-1,q)K^{xy}_{rdtw}(X_{p-1},Y_{q}) \\
   h(p-1,q-1) K^{xy}_{rdtw}(X_{p-1},Y_{q-1})  \\
    h(p,q-1)K^{xy}_{rdtw}(X_{p},Y_{q-1}) \\
   \end{array}
   \right.\\
\\
   K^{xx}_{rdtw}(X_{p},Y_{q}) = \frac{1}{3} \\
   \sum \left\{
   \begin{array}{ll}
    h(p-1,q) K^{xx}_{rdtw}(X_{p-1},Y_{q})e^{-\nu d_{E}^{2}(x(p),y(p))}  \\
    \Delta_{p,q} h(p,q)K^{xx}_{rdtw}(X_{p-1},Y_{q-1})e^{-\nu d_{E}^{2}(x(p),y(q))}   \\
    h(p,q-1)K^{xx}_{rdtw}(X_{p},Y_{q-1})e^{-\nu d_{E}^{2}(x(q),y(q))} \\
   \end{array}
   \right.\\
  \end{array}
\end{align}
where $\Delta_{p,q}$ is the Kronecker's symbol, $\nu \in \mathbb{R}^{+}$ is a \textit{stiffness} parameter which weights the local contributions, i.e. the distances between locally aligned positions, and $d_E(.,.)$ is a distance defined on $\mathbb{R}^{k}$. 

The initialization is simply $K^{xy}_{rdtw}(X_{0},Y_{0}) = K^{xx}_{rdtw} (X_{0},Y_{0}) = 1$.\\

The main idea behind this line of regularization is to replace the operators $\min$ and $\max$ (which prevent the symmetrization of the kernel) by a summation operator ($\sum$). This leads to consider, not only the best possible alignment, but also all the best (or nearly the best) paths by summing up their overall cost. The parameter $\nu$ is used to control what we call nearly-the-best alignment, thus penalizing more or less alignments too far from the optimal ones. This parameter can be easily optimized through a cross-validation. \\

\textbf{Elastic kernels}: we consider in this paper only the exponential kernel (Gaussian or RBF-type) constructed from the two previous elastic measures $d_{dtw}$ and $K_{rdtw}$,  and the non-elastic kernel obtained from the Euclidean distance \footnote {The Euclidean distance is usable only because a fixed number of skeletal positions is considered to characterize each movement, and this, irrespectively of their initial length}, i.e. $K_{dtw}(.,.)= e^{- d_{dtw}(.,.)/\sigma}$, and $K_{E}(.,.)= e^{- d_{E}^2(.,.)/\sigma}$. For the regularized DTW kernel, a data dependent normalization heuristic is  required and the final kernel takes the form:\\

$K_{rdtw}(.,.)= e^{\beta\mathcal{K}_{rdtw}^\alpha(.,.)/\sigma}$, with \\
\begin{itemize} 
\item $\alpha=1/log(max(\mathcal{K}_{rdtw}(.,.))/min(\mathcal{K}_{rdtw}(.,.)))$ and 
\item $\beta=exp(-\alpha \cdot log(min(\mathcal{K}_{rdtw}(.,.))))$,
\end{itemize}
 where $min$ and $max$ are taken over all the training data pairs.

%
%

\section{Experimentation}


To estimate the robustness of the proposed approach, we evaluate it on two motion capture databases  of opposite quality, the first one developed at the Max Planck Institute, the other at Microsoft research laboratories. \\

\textbf{HDM05 data set} \cite{HDM05} consists of data captured at 120hz by a Vicon MX system composed of a set of reflective optical markers followed by six high-definition cameras and configured to record data at 120hz. 
The movement sequences are segmented and transformed into sequences of skeletal poses consisting of N = $31$ joints, each associated to a 3D position $(x, y, z)$. In practice the position of the root of the skeleton (located near its center of mass) and its orientation serving as referential coordinates, only the relative positions of the remaining 30 joints are used, which leads to represent each position by a vector $Y_T \in \mathbb{ R}^{k}$ , with $k=90$. We consider two recognition/classification tasks:  HDM05-1 and HDM05-2 that are respectively those proposed in \cite{OfliF2012} (also exploited in the work of \cite{Hussein2013}) and \cite{OfliF2013}. For both tasks, three subjects are involved during learning and two separate subjects are involved during testing. For task HDM05-1, 11 gestural actions are processed: \textit{\{deposit floor, elbow to knee, grab high, hop both legs, jog, kick forward, lie down floor, rotate both arms backward, sneak, squat, and throw basketball\}}. This constitutes 249 motion sequences. For task HDM05-2 , the subjects are the same, but five additional gestural actions are considered in addition to the previous 11: \textit{\{jump, jumping jacks, throw, sit down, and stand up\}}. For this task, the data set includes 393 movement sequences in total.
For both tests, the lengths of the gestural sequences are between 56 and 901 postures (corresponding to a movement duration between 0.5-7.5 sec.) . \\
 
\textbf{MSR-Action3D data set}: This database \cite{Li2010} has recently been developed to provide a Kinect data \textit{benchmark}. It consists of 3D depth image sequences (\textit{depth map}) captured by the Microsoft Kinect sensor. It
contains 20 typical interaction gestures with a game console that are labeled as follows \textit{high arm wave, horizontal arm wave, hammer, hand catch, forward punch, high throw, draw x, draw tick, draw circle, hand clap, two hand wave, side-boxing, bend, forward kick, side kick, jogging, tennis swing, tennis serve, golf swing,  pickup \& throw} . Each action was carried out by 10 subjects facing the camera, 2 or 3 times. This data set includes 567 motion sequences whose lengths vary from 14 to 76 skeletal poses. The 3D images of size $640 \times 480$  were 
captured at a frequency of 15hz.  From each 3D image a skeletal posture has been extracted with $N=20$ joints, each one being characterized by three coordinates. As for the previous data set, we characterize postures relatively to the referential coordinates located at the root of the skeleton, which leads to represent each posture by a vector $Y_t \in \mathbb{R}^{k}$, with $k=3 \times 19 = 57$.
The task is to provide a cross-validation on the subjects, i.e. 5 subjects participating in learning and 5 subjects participating in testing, considering all possible configurations which represent 252 learning/testing pairs in total.

\subsection{Results and analysis}

For the two considered tasks, we present the results obtained using a SVM classifier built from the LIBSVM library \cite{Libsvm01}, the elastic kernels $K_ {dtw}$ and $K_{rdtw}$, and as a baseline the Euclidean distance kernel, $K_ {E}$. 

\begin{figure}[ht]
  \centering
  \begin{tabular}{cc}
	\includegraphics[width=75mm]{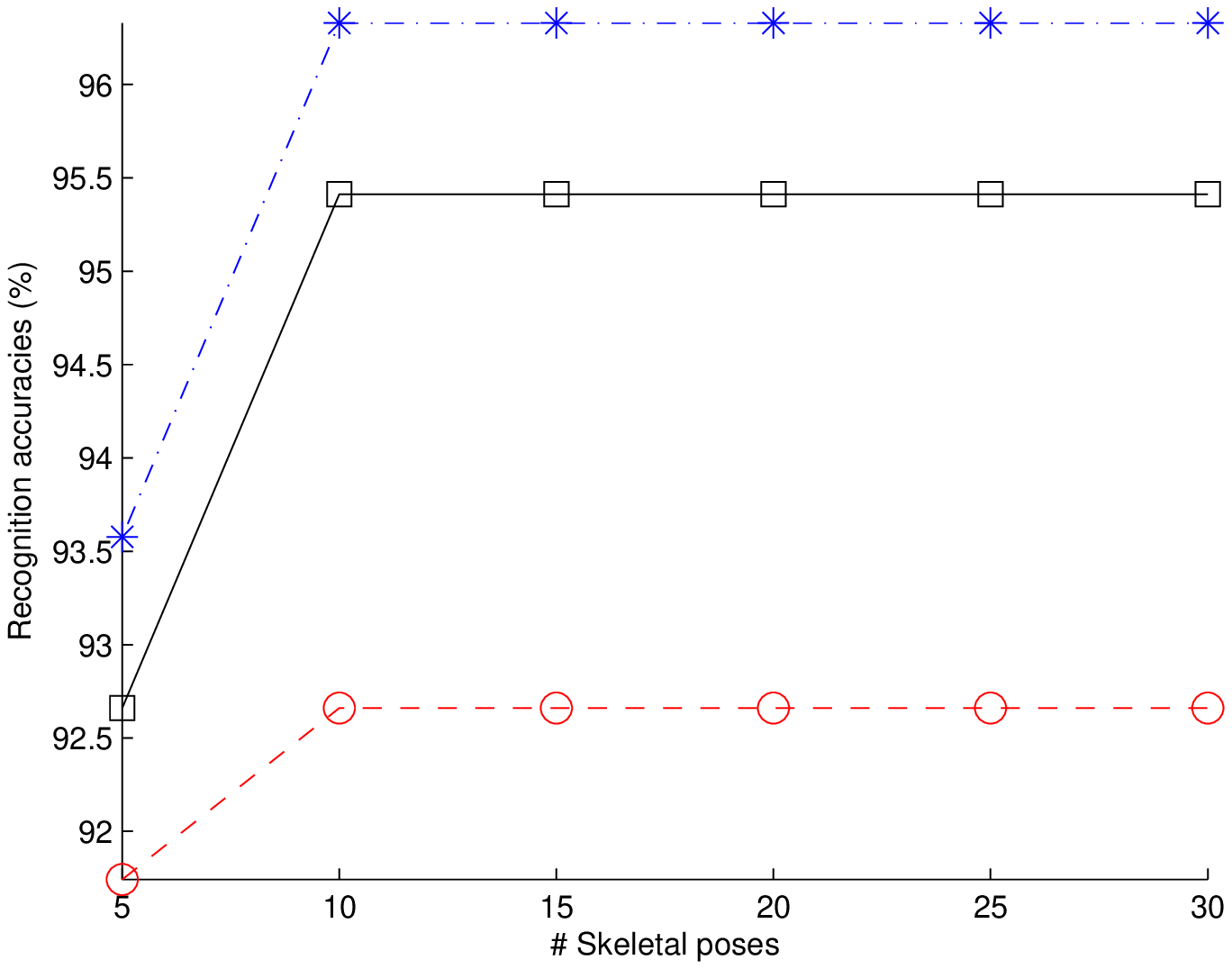} \\
    \includegraphics[width=75mm]{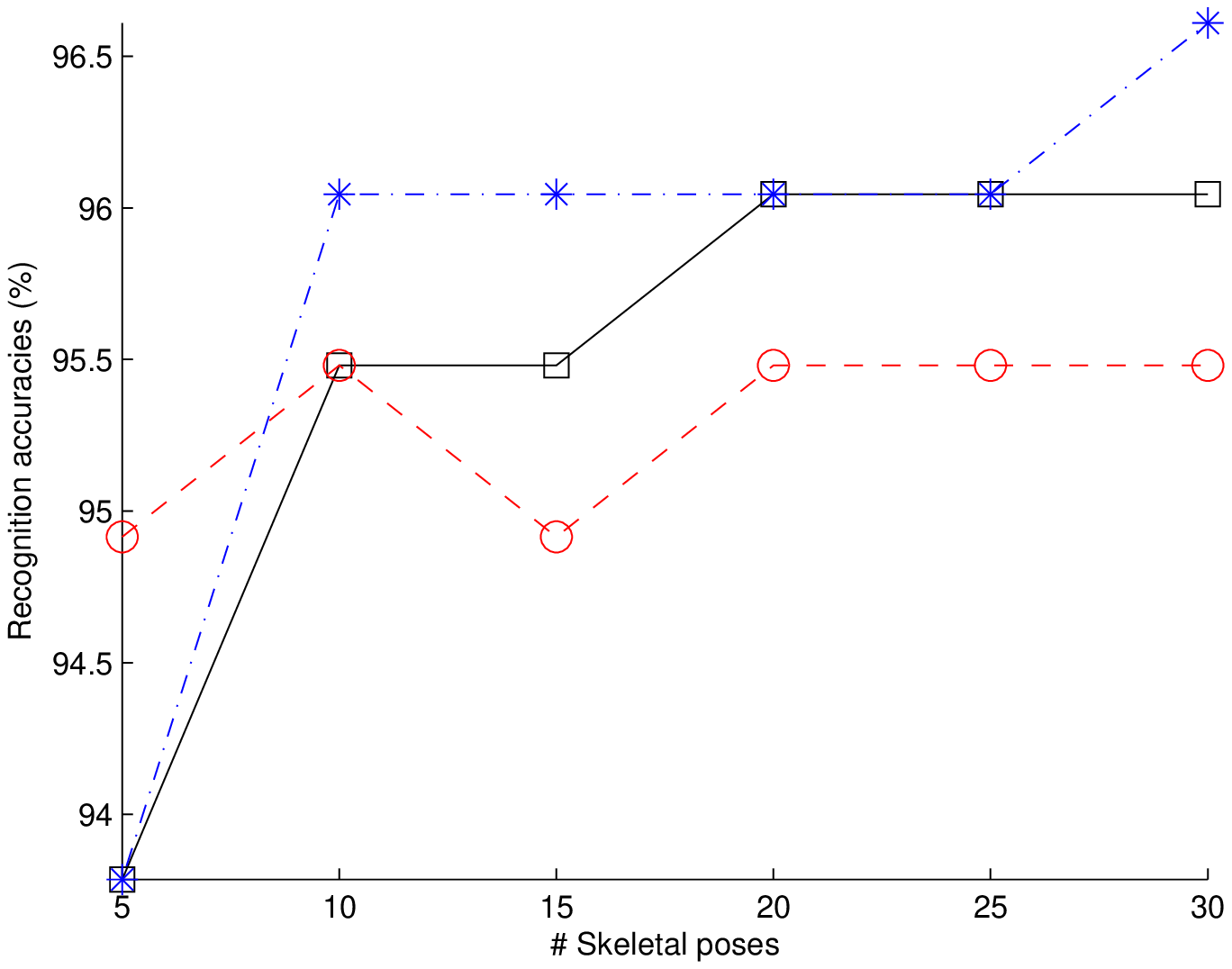}
  \end{tabular}
  \caption{Classification accuracies for HDM05-1 task 
   (top), and HDM05-2 task 
  (bottom), when the number of skeletal poses varies: $K_{E}$ (red, circle, dash),  $K_{dtw}$ (black, square, plain), $K_{rdtw}$ (blue, star, dotted). }
\label{fig:hdm05}
\end{figure}

\begin{figure}[ht]
  \centering
  \begin{tabular}{cc}
	\includegraphics[width=75mm]{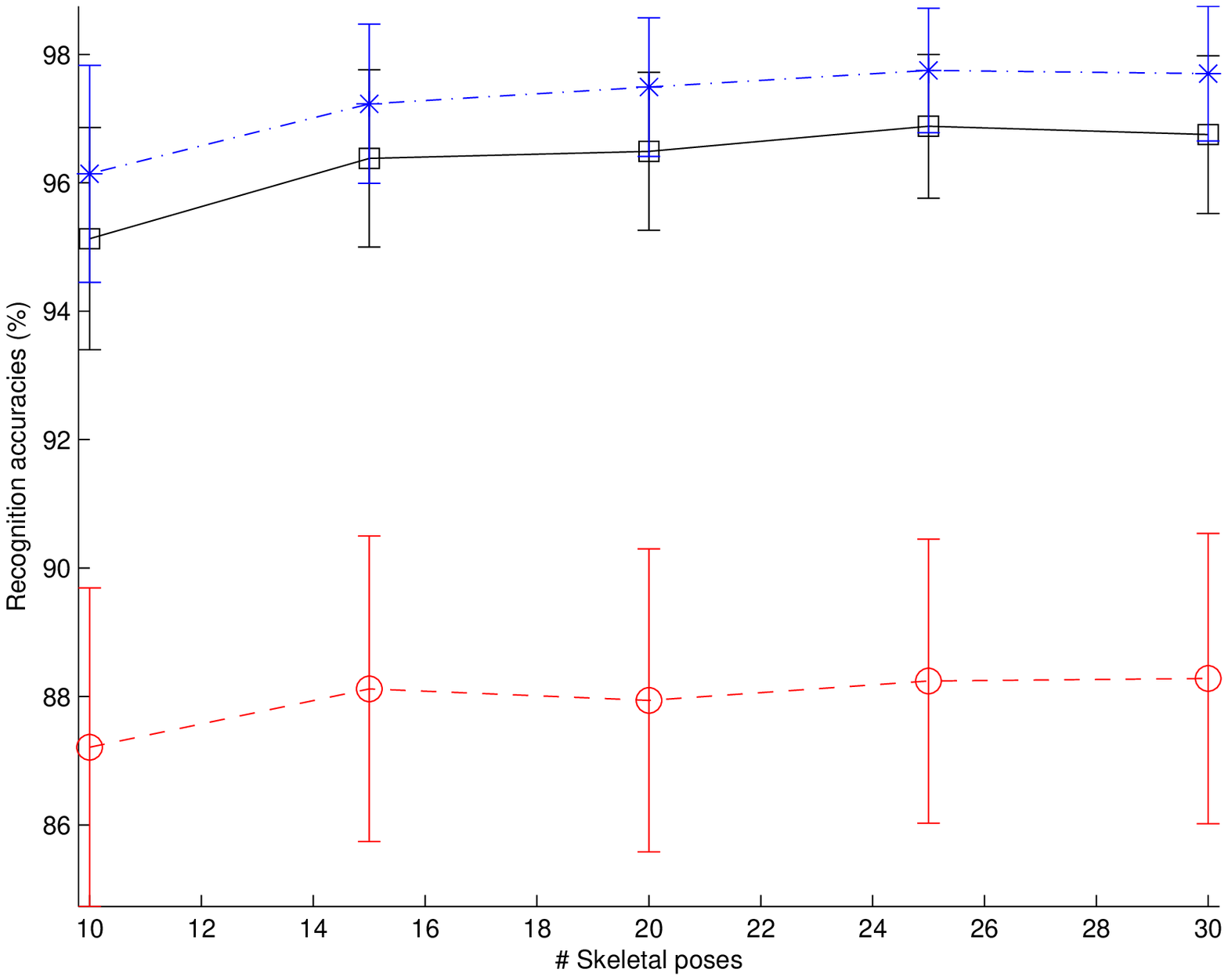}\\
    \includegraphics[width=75mm]{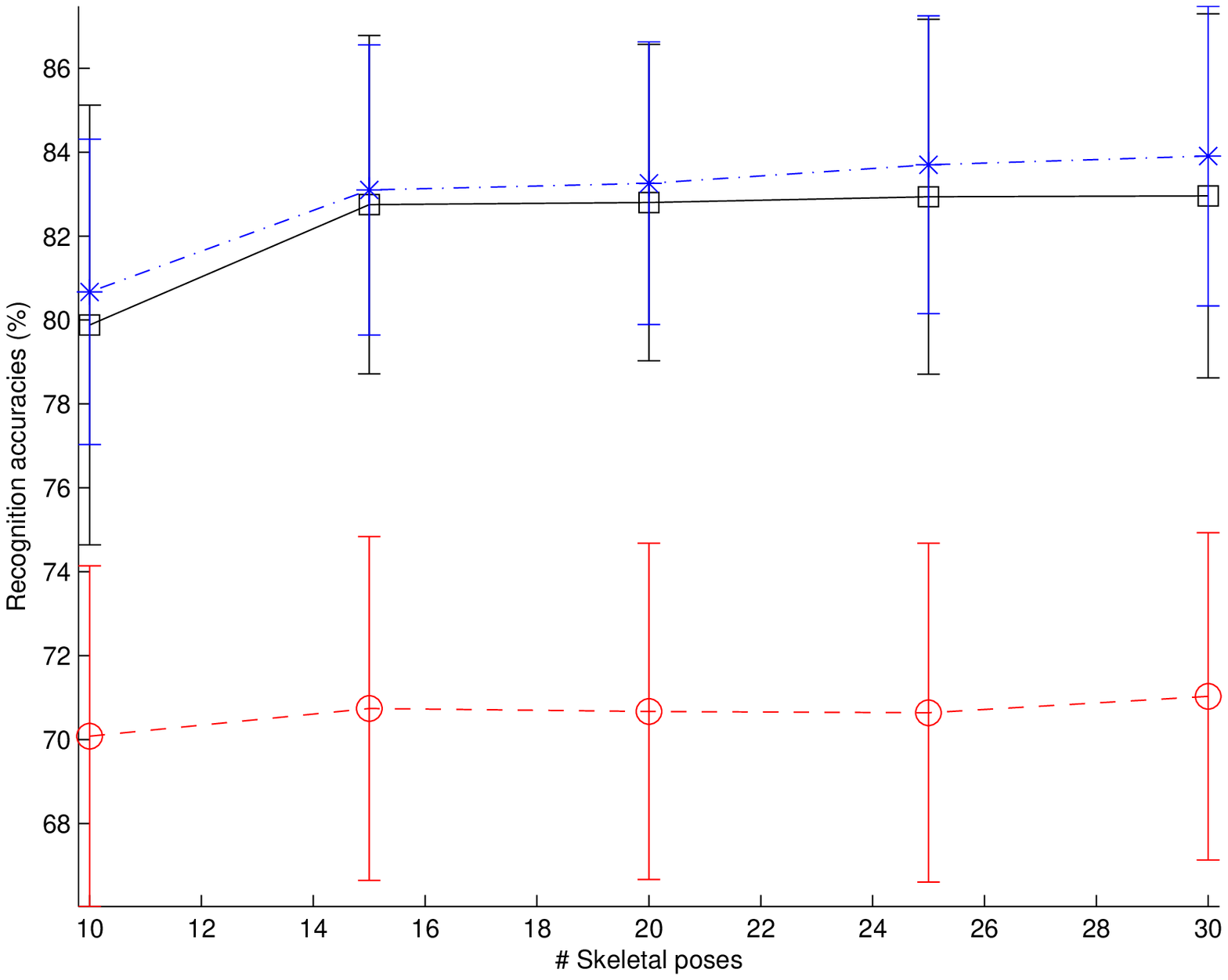}
  \end{tabular}
  \caption{Classification accuracies for the  MSRAction3D data set, on the learning data (top) and testing data (bottom) when the number of skeletal poses varies: $K_{E}$ (red, circle, dash),  $K_{dtw}$ (black, square, plain), $K_{rdtw}$ (blue, star, dotted). }
\label{fig:MSR}
\end{figure}

Figure \ref{fig:hdm05} presents the classification accuracies when the number of skeletal postures selected after downsampling varies between 5 and 30.    
Results for HDM05-1 task is presented in the top sub-figure, while results for HDM05-2 task is given in the bottom sub-figure. Figure \ref{fig:MSR} shows also the classification accuracies obtained on the MSRAction3D data set when the number of skeletal postures varies between 10 and 30: the accuracies obtained for a 10-fold cross validation on the training data is given in the top sub-figure.  accuracies for the testing data is given in the bottom sub-figure. 

On both figures, we observe that the sub-sampling does not catastrophically degrade the accuracies. High levels of down-sampling (e.g. 10 to 15 postures retained by movement, which represents an average compression ratio of 97 \% for  HDM05 and 70 \% on MSRAction3D) lead to very satisfactory results (96-98 \% for the two HDM05 tasks and 95 to 97 \% for the MSRAction3D task on the learning data). The SVM classifier constructed on the basis of the regularized kernel $K_{rdtw }$  produces the best recognition rate. We note that the MSRAction3D task is more difficult:  much lower performance are obtained for the SVM built on the basis of the Euclidean distance; in addition, if very good classification rate (96 \%) is obtained on the training data, due to the noisy nature of Kinect data and the inter subject variability, the recognition rate on the test data drop down to 82 \% .

\begin{table*}[!ht]
 \begin{center}
   \tabcolsep = 2\tabcolsep
   \begin{tabular}{lcccccc}
   \hline\hline
                & $K_{E}$ A & $K_{E}$ T &   $K_{dtw}$ A & $K_{dtw}$ T & $K_{rdtw}$ A & $K_{rdtw}$ T\\
   \hline
   Mean & 87,71	& 69,73 & 96,04	& 81,41	& 96,65	& 82,50        \\
   Stand. dev. & 2,34	& 5,73 & 1,36 & 5,04 & 1,13	 &3,22	 \\
   \hline
   \end{tabular}
\caption{Means and standard deviations of classification accuracies on the MSRAction3D data set obtained according to a cross-validation on the subjects (252 tests) A: on the training data, T: on test data for a number of skeletal postures equal to 15.} \label{tab:resMSRAction3D}
 \end{center}
\end{table*}

Table \ref{tab:resMSRAction3D} gives for the MSRAction3D data set and for the SVM based on $K_{E}, K_{dtw}$ and $K_{rdtw}$ kernels, means and standard deviations, obtained on the training data (A) and testing data (T), of recognition rates (classification accuracies)  when performing the cross-validation over the 10 subjects (252 configurations). For this test, movements are represented as sequences of 15 skeletal postures. The drop of accuracies between Learning and testing is due, on this dataset,  to the large inter subjects variability of movement realizations.

For comparison, table \ref{tab:resComp} gives results obtained by different methods of the state-of-the-art and compare them with the performance of our SVM constructed from the Regularized DTW associated with a down-sampling of 15 postures. To that end, we have reimplemented the Cov3DJ approach \cite{Hussein2013} to get, for the MSRAction3D data set, the average result given by a 5-5 cross-validation on the subjects (252 tests). This comparative analysis shows that the SVM constructed from regularized DTW kernel provides results slightly above the current state-of-the-art for the considered data sets and tasks.

\begin{table}[ht]
 \begin{center}
   \tabcolsep = 2\tabcolsep
   \begin{tabular}{lc}
   \hline\hline
  HDM05-1 &  Accuracy (\%)\\
   \hline
SMIJ \cite{OfliF2012} & 84.40\\
Cov3DJ, L = 3 \cite{Hussein2013} & 95.41\\
$SVM K_{rdtw}$, 15 poses & \textbf{96.33} \\
   \hline
   \hline
     HDM05-2 &  Accuracy (\%)\\
   \hline
SMIJ \cite{OfliF2013}, 1-NN & 91.53 \\
SMIJ \cite{OfliF2013}, SVM & 89.27 \\
$SVM K_{rdtw}$, 15 poses & \textbf{96.05}\\
\hline\hline
MSR-Action3D &  Accuracy (\%)\\
\hline
Cov3DJ, L=3 \cite{Hussein2013} & $72.33 \pm 3.69$ \footnotemark[2]\\
HON4D, \cite{Oreifej2013}, & $82.15 \pm 4.18$\\
$SVM K_{rdtw}$ 15 poses, & \textbf{83.10 $\pm$ 3.46}\\
\hline
   \end{tabular}
\caption{Comparative study.} 
\label{tab:resComp}
 \end{center}
\end{table}
\footnotetext[2]{according to our own implementation of Cov3DJ}

\begin{figure}[ht]
  \centering
  \begin{tabular}{cc}
	\includegraphics[width=75mm]{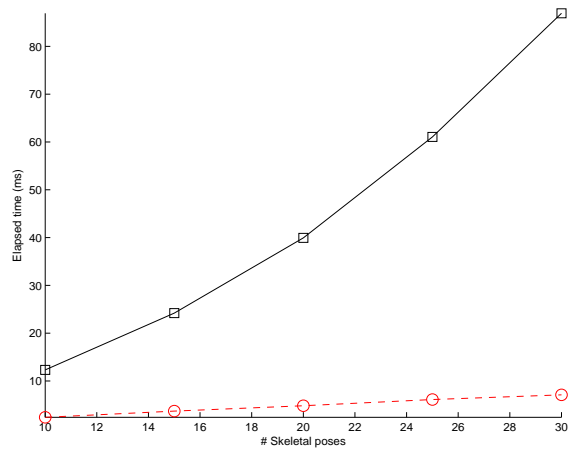}\\
  \end{tabular}
  \caption{Elapsed time as a function of the number of skeletal poses (10 to 30 poses): i) Euclidean Kernel, Red/round/dotted line, ii) Elastic kernel (RDTW), Black/square/plain line. }
\label{fig:elapsedTime}
\end{figure}
Finally, in Figure \ref{fig:elapsedTime}, we give the average CPU elapsed time for the processing of a single gestural MSRAction3D action when varying the number of retained skeletal poses. The test has been performed on an Intel Core i7-4800MQ CPU, 2.70GHz. Although the computational cost for the elastic kernel is quadratic, the latency for the classification of a single gestural action is less than 25 milliseconds when 15 poses are considered, which effectively meets easily \textit{real-time} requirements. 

\section{Conclusion and perspectives}

In the context of isolated gesture recognition, where few studies explicitly consider dimension reduction along the time axis, we have presented a simple approach based on sub-sampling motion sequences coupled to the exploitation of elastic kernel machines. On the data sets and tasks that we have addressed, we have shown that, even when quite important down-sampling is considered,  the recognition accuracy only slightly degrades. The temporal redundancy is therefore high and apparently not critical for the discrimination of the selected movements and tasks. In return, the down-sampling benefits in terms of computational complexity is quadratic with the reduction of the number of skeletal postures kept along the time axis. 

Furthermore, the elasticity of the kernel provides a performance gain (compared to kernel based on the Euclidean distance) which is important when the data are characterized by high variability. Our results show that a SVM based on a regularized DTW kernel is very competitive comparatively to the state-of-the-art methods applied on the two tested data sets, even when the dimension reduction on the time axis is important. This study opens perspectives to the use of more sophisticated elastic kernels \cite{MarteauTWED09} associated to adaptive sampling techniques \cite{Marteau05} \cite{MarteauPAA09} capable of extracting the most significant and discriminant skeletal poses in movement sequences. 



\bibliographystyle{IEEEtran}


\end{document}